\documentclass[11pt]{article}
\usepackage[a4paper,margin=1in]{geometry}
\usepackage[T1]{fontenc}
\usepackage[utf8]{inputenc}
\usepackage{times}
\usepackage{microtype}
\usepackage{graphicx}
\usepackage{booktabs}
\usepackage{array}
\usepackage{multirow}
\usepackage{amsmath,amssymb}
\usepackage{pifont}
\usepackage{xcolor}
\usepackage{enumitem}
\usepackage[font=small,labelfont=bf]{caption}
\usepackage[numbers,sort&compress]{natbib}
\usepackage[colorlinks=true,linkcolor=blue!50!black,citecolor=blue!50!black,urlcolor=blue!50!black]{hyperref}
\setlist{itemsep=1pt,topsep=3pt}
\newcommand{\mk}[1]{\texttt{\small #1}}

\title{\textbf{Escriba-OCR: A 1B Vision-Language Model for Error-Preserving Transcription of Handwritten Portuguese Essays}}
\author{Tardelli Ronan Coelho Stekel\\[2pt]
  \small Federal Institute of São Paulo (IFSP), Brazil}
\date{November 2026}

\begin{document}
\maketitle

\begin{abstract}
\noindent
Feedback tools and research on learner writing need transcriptions of handwritten essays that keep exactly what the student wrote: misspellings, line breaks, struck-through words and insertions. Paleographers call this a \emph{diplomatic} transcription. Vision-language models (VLMs) read handwriting well, but they also tend to rewrite it into correct language. We measure this behaviour on BRESSAY, a dataset of 1,000 handwritten Brazilian Portuguese essays, with a benchmark that adds error preservation, invented errors, edit-markup precision and recall and line structure to the character error rate (CER), and that compares systems with paired bootstrap confidence intervals. Because schools need models they can run on their own hardware, we restrict the study to open-weight systems.

We evaluate 22 open baselines: 16 general-purpose VLMs from 2B to 400B parameters and five OCR-specialised VLMs, all without task-specific training, and a recogniser trained on BRESSAY lines. The best of them, Qwen3-VL 235B-A22B, reaches 5.6\% CER on the 199 test pages. None reproduces the edit markup reliably, and every VLM replaces 39--78\% of the words that are absent from a Portuguese dictionary with dictionary words, including general VLMs whose prompt explicitly asks them not to.

We then fine-tune LightOnOCR-3-1B on the 647 training pages in 85 minutes on one GPU. The resulting model, Escriba-OCR-1B, reaches 2.06\% CER, 3.5 points below the best zero-shot baseline (95\% CI $[-4.2,-2.9]$), with 1.63\% CER on the text alone. Under the case-sensitive metric of the ICDAR 2024 BRESSAY competition it reaches 2.1\% CER, below the best published page-level results (2.2--2.4\%). It recovers 77\% of illegible deletions and 67\% of insertions and reproduces the line structure of 97\% of the pages, but it still replaces 60\% of the likely misspellings with dictionary words. Three training seeds agree to within 0.04 CER points, and quantised to 4 bits the model runs on an ordinary CPU at about one minute per page, with CER within 0.1 points of the GPU model. On 40 letters from another handwriting corpus it reads as well as its base model overall (5.1\% against 4.9\% CER once the base model's appended comments are removed), but misreads words that never occur in BRESSAY, such as first-person verbs and proper nouns, more often (40\% against 30\%); interpolating its weights with the base model's recovers about half of this gap and lowers BRESSAY test CER to 2.00\%, at the cost of edit-markup recall.

A teacher's review of 96 apparent corrections finds that 45\% of the 89 judgeable ones (95\% CI 35--55\%) are errors in the BRESSAY ground truth, so dictionary-based scores likely understate error preservation. We report what did not reduce normalisation (contrastive decoding against the text-only prior, over-sampling misspelled words, higher input resolution), analyse error types, per-page behaviour and data scaling, and release the model, the benchmark code and the review data.
\end{abstract}

\section{Introduction}

Handwritten essays are the norm in Brazilian schools and in the ENEM, the national university-entrance exam, whose essay (\emph{redação}) millions of students write every year. Automatic transcription would let teachers use digital feedback tools and let researchers study learner writing at scale, and pipelines that score ENEM essays already digitise handwriting with language-model-based OCR followed by text normalisation \citep{rossman2026enem}. For feedback and research, however, a clean text is the wrong target. A teacher giving spelling feedback needs \emph{caza} where the student wrote \emph{caza}, not the corrected \emph{casa}, and a researcher counting orthographic errors needs every one of them.

Paleography calls a transcription that reproduces a document exactly as written a \emph{diplomatic} transcription: original spelling and mistakes, line breaks and end-of-line hyphenation, deletions and insertions. Its opposite is a \emph{normalised} transcription, which corrects and modernises the text. In this paper we use \emph{error-preserving transcription} for the diplomatic case.

Modern VLMs read handwriting remarkably well \citep{crosilla2025benchmarking,humphries2024unlocking}, but they are also strong language models, and strong language models normalise. Recent work shows that VLM-based OCR rewrites perturbed or implausible text toward plausible forms \citep{lee2026readorrewrite,liang2026deepseekocr,karamolegkou2026readingorguessing}, that many handwriting errors of multimodal LLMs come from language priors \citep{zhang2026wildhandbench}, and that a low CER can hide orthographic normalisation \citep{gardella2026lowcer}. Learner text is where this matters most, because its ``errors'' are the signal. To our knowledge, the extent of the problem has not been measured on handwritten learner essays in Portuguese, and no error-preserving recogniser is available for them.

Schools also face a practical constraint. Student essays are personal data under Brazil's data-protection law (LGPD), so a transcription system should run on hardware the school controls. We therefore restrict this study to open-weight models, which can be audited, fine-tuned and deployed locally; proprietary APIs are outside its scope. This restriction need not cost quality: on a native benchmark of Brazilian Portuguese text embeddings, an openly licensed, self-hostable model reaches the leading tier \citep{stekel2026mtebbr}, and we ask whether the same holds for transcription.

\begin{figure}[t]
\centering
\includegraphics[width=\textwidth]{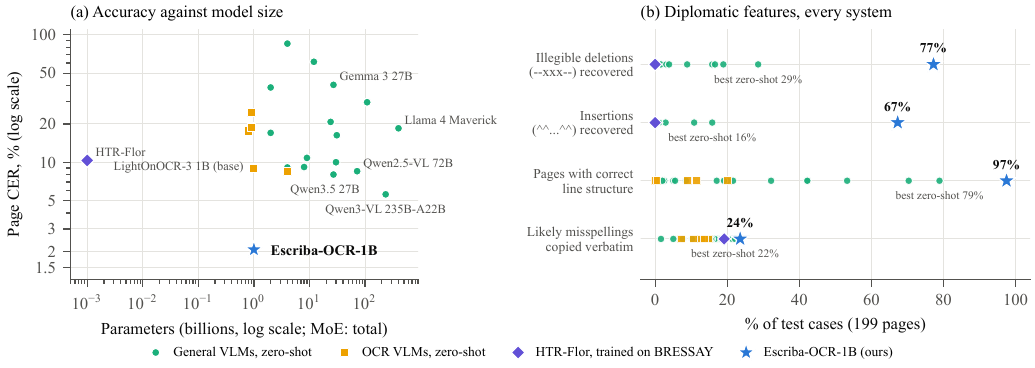}
\caption{Escriba-OCR-1B against the 22 open baselines on the 199 BRESSAY test pages. (a)~Page CER against model size; MoE models are plotted at their total parameter count. Fine-tuning puts the 1B model far below zero-shot VLMs up to 400 times larger, including its own base model. (b)~The features of a diplomatic transcription, one dot per system. The best zero-shot deletion recall (29\%) belongs to Gemma 3 4B at 0.5\% precision. HTR-Flor reads ground-truth line crops, so its line structure is not shown. Error preservation, the last row, remains the open problem for every system (§\ref{sec:review}).}
\label{fig:overview}
\end{figure}

We study the problem on BRESSAY \citep{neto2024bressay}, 1,000 handwritten Brazilian Portuguese essays transcribed with their errors and edit markup, and make four contributions:
\begin{enumerate}
\item a benchmark for error-preserving transcription that reports page-level CER and WER, error preservation and invented errors, edit-markup precision and recall and line structure, with paired bootstrap confidence intervals (§\ref{sec:metrics});
\item an evaluation of 22 open baselines, 21 VLMs from 0.8B to 400B parameters used without task-specific training and a trained HTR model, which shows that none reproduces the markup reliably and that the VLMs normalise most misspellings (§\ref{sec:results});
\item Escriba-OCR-1B, a 1B model fine-tuned in 85 minutes that reduces CER by a factor of 2.7 relative to the best zero-shot baseline, improves on the best published page-level BRESSAY result under our reconstruction of the competition's metric, reproduces edit markup and runs on a single GPU or, quantised, on a CPU (Figure~\ref{fig:overview}; §\ref{sec:escriba}, §\ref{sec:results});
\item an analysis of what helps and why, including error decomposition, per-page behaviour, data scaling, negative results on decoding and sampling remedies, a human review showing that the BRESSAY ground truth itself explains almost half of the apparent corrections, and an out-of-domain test on a second corpus with weight interpolation as a remedy for the vocabulary that fine-tuning narrows (§\ref{sec:analysis}).
\end{enumerate}

\section{Related Work}

\paragraph{Handwritten text recognition.} Line-level CRNN/CTC models such as HTR-Flor \citep{neto2020htrflor}, trained on benchmarks like IAM \citep{marti2002iam}, were the standard until page-level, segmentation-free models such as DAN \citep{coquenet2023dan} and transformer recognisers such as TrOCR \citep{li2023trocr}. BRESSAY \citep{neto2024bressay} was the basis of an ICDAR 2024 competition \citep{neto2024competition}, won by the PERO system from Brno \citep{kodym2021pero,kohut2023finetuning} with 3.77\% page CER, and segmentation-free models of the DAN family later reported 2.33--2.43\% in their main results and 2.24\% in their best ablation setting \citep{metadan2025}. \citet{neto2026htsr} train four line recognisers on BRESSAY with handwriting synthesised by a conditional GAN; averaged over the four, line CER on the full training set is about 12\% (11.5--12.2\% across their training regimes, and 6.4\% for their best model), and they note that none of their models was designed for the edit markup. All of these systems are trained on the BRESSAY training split, and Table~\ref{tab:published} compares Escriba with them under the competition's metric. Earlier work by the BRESSAY authors applied language-model spelling correction to HTR output \citep{neto2020spelling}, the opposite of our goal, and LHTR.br \citep{silva2023lhtrbr} is another Brazilian Portuguese handwriting dataset.

\paragraph{OCR with vision-language models.} End-to-end document models \citep{kim2022donut,blecher2023nougat} and VLM-based OCR systems \citep{poznanski2025olmocr,lightonocr2026,cui2025paddleocrvl,duan2026glmocr} transcribe whole pages, mostly printed ones, into Markdown, and general VLMs \citep{bai2025qwen25vl,bai2025qwen3vl,gemma2025gemma3} read handwriting when prompted. Zero-shot evaluations of multimodal LLMs on handwriting \citep{crosilla2025benchmarking,humphries2024unlocking,boillet2026metatr} report strong accuracy on historical and modern documents. A growing body of work studies \emph{faithfulness}: VLMs ``read or rewrite'' \citep{lee2026readorrewrite}, lean on linguistic crutches \citep{liang2026deepseekocr}, lose visual grounding on unusual text \citep{karamolegkou2026readingorguessing} and hallucinate plausible words behind a low CER \citep{gardella2026lowcer}. We bring this question to learner writing, where unfaithfulness erases exactly the information users need.

\paragraph{Learner handwriting.} \citet{gold2023preserving} proposed annotation guidelines for transcripts that retain orthographic features, with double transcription and high agreement. \citet{gold2023learner} re-transcribed 1,350 pages of German learner writing accordingly and showed that the decoding dictionary strongly affects whether errors survive. \citet{yasin2026showdonttell} study visual error feedback on handwriting, find that neither modular nor end-to-end systems yet reach acceptable quality, and note that recognisers over-correct learner errors. We follow this line for Brazilian Portuguese, measure how modern open VLMs behave, and release a model trained for the task.

\section{Data}
\label{sec:data}

BRESSAY \citep{neto2024bressay} contains 1,000 pages of Brazilian Portuguese argumentative essays, one writer per page, collected from online essay platforms and transcribed line by line with the student's spelling. Edits are marked as \mk{--text--} (struck through, legible), \mk{--xxx--} (struck through, illegible), \mk{\#\#text\#\#} and \mk{\$\$text\$\$} (inserted above or below the line) and \mk{@@???@@} (illegible). We use the official page-level splits (Table~\ref{tab:data}).

\begin{table}[ht]
\centering\small
\resizebox{\textwidth}{!}{%
\begin{tabular}{lrrrrrrrr}
\toprule
Split & Pages & Lines & Words & Chars/page & Image (px) & px/line & \mk{--t--} / \mk{--xxx--} / \mk{\#\#t\#\#} & Non-dict. words \\
\midrule
Train & 647 & 19,565 & 267,588 & 2,712 & 492$\times$556 & 18.4 & 1,823 / 902 / 930 & 1,820 \\
Validation & 154 & 4,609 & 63,957 & 2,724 & 525$\times$584 & 19.5 & 376 / 184 / 140 & 382 \\
Test & 199 & 5,916 & 81,118 & 2,669 & 538$\times$618 & 20.8 & 590 / 259 / 379 & 496 \\
\bottomrule
\end{tabular}}
\caption{BRESSAY statistics. Image sizes are medians; px/line is the median image height divided by the mean number of lines per page. A page holds about 30 lines and 410 words. The last column counts reference words absent from a pt-BR dictionary (§\ref{sec:metrics}).}
\label{tab:data}
\end{table}

\noindent Three properties of these data shape our results. Pages are cropped to the text, so a handwritten line is only about 19\,px tall, and OCR-specialised processors such as LightOnOCR's only downscale images, so upscaling is necessary. ALL-CAPS handwriting is transcribed in sentence case, a convention that a fine-tuned model can learn and a prompted model cannot. Finally, the transcriptions themselves contain mistakes, which we quantify in §\ref{sec:review}. Model selection for Escriba used only the validation split, and the test split was used only to evaluate the selected configuration. For the zero-shot baselines, image upscaling and a repetition penalty were set after an early test run showed repetition loops (§\ref{sec:systems}). BRESSAY's terms restrict its use to non-commercial research and teaching.

\section{Benchmark and Metrics}
\label{sec:metrics}

For page $p$ with reference $r_p$ and hypothesis $h_p$, both are first normalised: Unicode NFC, visually identical glyphs merged (curly to straight quotes, \mk{°} to \mk{º}, dashes to \mk{-}), whitespace collapsed. The micro-averaged character error rate is then
\begin{equation}
\mathrm{CER} = \frac{\sum_p \min\big(d(r_p, h_p),\, |r_p|\big)}{\sum_p |r_p|},
\label{eq:cer}
\end{equation}
where $d$ is the Levenshtein distance. Each page is capped at 100\% so that a single runaway generation, such as a repetition loop, cannot dominate the average. We report three variants, all capped: CER (case-insensitive, markup included, our main metric), CER$_\text{case}$ (case-sensitive) and CER$_\text{text}$ (case-insensitive, with struck-through and illegible spans removed and insertions unwrapped, which measures reading of the text alone). Markup is parsed line by line, as in BRESSAY, so an unclosed marker in a model's output affects only its own line. WER is computed in the same way at the word level, without the cap.

\paragraph{Handling of student errors.} Let $\mathcal{D}$ be the Hunspell pt-BR dictionary. After rejoining words split across lines and removing markup, reference words that are lowercase-initial, at least three letters long and not in $\mathcal{D}$ form the set $S$ of \emph{likely misspellings}; the test split has $|S| = 496$. With a word-level Levenshtein alignment $a(\cdot)$ between reference and hypothesis, we report
\begin{align}
\mathrm{Pres} &= \frac{\left|\{w \in S : a(w) = w\}\right|}{|S|}, &
\mathrm{Corr} &= \frac{\left|\{w \in S : a(w) \neq w,\ a(w) \in \mathcal{D}\}\right|}{|S|},
\end{align}
together with the shares of words \emph{misread} (replaced by another non-word) and \emph{deleted}. Conversely, the \emph{invented-error} rate is the share of reference dictionary words aligned to a non-dictionary word. Preservation alone would reward a recogniser that misreads everything, so we always report it together with invented errors. $S$ approximates the true misspellings: it also contains foreign words, proper nouns and transcription errors, which we quantify in §\ref{sec:review}. With about 2.5 such words per page, preservation has a 95\% interval of roughly $\pm4$ points.

\paragraph{Markup and structure.} Each marked span is reduced to a (type, content) pair, and precision and recall are computed on the multiset intersection of reference and hypothesis pairs per page. Matching ignores position, and for illegible deletions (\mk{--xxx--}) the content is empty, so for that type we report precision next to recall. \emph{Lines} is the share of pages whose output has exactly as many non-empty lines as the reference.

\paragraph{Significance.} We compare systems with a paired page-level bootstrap \citep{koehn2004bootstrap}: the 199 test pages are resampled with replacement $B = 2{,}000$ times, both systems' metrics are recomputed on each sample, and we report the 2.5th and 97.5th percentiles of the difference $\Delta$. The bootstrap captures variation over pages but not over training runs. From two decoder-only runs on validation that differ only in the seed we estimate seed-to-seed variation at about 0.6 CER points, and two vLLM evaluations of identical weights differed by about 0.5 points because of batched-inference non-determinism (one page looped in one of them). We therefore read differences below about 1 point between decoder-only variants as inconclusive. Training the full model, vision encoder included, is far more stable: three seeds differ by at most 0.04 CER points on the test set (§\ref{sec:results}), so we treat smaller differences between full-model variants as tentative rather than inconclusive.

\section{Systems}
\label{sec:systems}

All baselines are open-weight models evaluated without task-specific training, except HTR-Flor, which we train on BRESSAY. Table~\ref{tab:settings} summarises their settings.

\paragraph{General-purpose VLMs.} Thirteen instruction-tuned VLMs are queried through OpenRouter with an error-preserving instruction in Portuguese. The prompt asks the model to copy exactly, to leave spelling, accents, agreement, punctuation and casing uncorrected, to keep line breaks and end-of-line hyphens and to use the BRESSAY markup; the full prompt is released with the code. The models are Gemma 3 4B, 12B and 27B \citep{gemma2025gemma3}, Gemma 4 31B \citep{google2026gemma4}, Qwen2.5-VL 72B \citep{bai2025qwen25vl}, Qwen3-VL 8B, 30B-A3B and 235B-A22B \citep{bai2025qwen3vl}, Qwen3.5 9B and 27B \citep{qwen2026qwen35}, Llama 4 Scout and Maverick \citep{meta2025llama4} and Mistral Small 3.2 \citep{mistral2025small32}. All OpenRouter queries were made in November 2026. OpenRouter routes requests to third-party providers whose quantisation we cannot control; several models were served by two to five providers (Table~\ref{tab:providers} in the appendix). Three small VLMs, Qwen3.5 2B and 4B and Qwen3-VL 2B, run locally with vLLM \citep{kwon2023vllm} and the same prompt.

\paragraph{OCR-specialised VLMs.} LightOnOCR-3 0.8B, 1B and 4B \citep{lightonocr2026,lightonocr3blog2026}, PaddleOCR-VL 1.6 \citep{cui2025paddleocrvl,zhang2026paddleocrvl16} and GLM-OCR \citep{duan2026glmocr} run locally with their native prompts, since they do not follow instructions. Local models decode greedily with a repetition penalty of 1.05. These settings matter: without upscaling and the penalty, Qwen3.5 2B looped on 106 of the 199 test pages.

\paragraph{HTR.} HTR-Flor \citep{neto2020htrflor} is trained on the BRESSAY training lines for 60 epochs with the authors' code and evaluated on ground-truth line crops, which gives it perfect segmentation, an advantage over page-level systems. Its lines are assembled into pages for scoring. It produced no markup at all, although the training targets contain it; we did not investigate whether the toolkit filters these characters.

\begin{table}[t]
\centering\small
\resizebox{\textwidth}{!}{%
\begin{tabular}{llllll}
\toprule
Systems & Run via & Input & Prompt & Decoding & Max new tokens \\
\midrule
Escriba-OCR-1B & vLLM 0.34, A100 & 1200\,px PNG & image only & greedy, rep.\ penalty 1.05 & 2,048 \\
13 general VLMs & OpenRouter & 1540\,px JPEG (q92) & error-preserving instruction & temperature 0, reasoning off & 3,000 \\
3 small general VLMs & vLLM 0.34, L4 & 1600\,px PNG & error-preserving instruction & greedy, rep.\ penalty 1.05 & 2,048 \\
5 OCR-specialised VLMs & vLLM 0.34, L4 & 1600\,px PNG & native (empty or task tag) & greedy, rep.\ penalty 1.05 & 2,048 \\
HTR-Flor & authors' toolkit & GT line crops & -- & CTC beam search (width 32) & -- \\
\bottomrule
\end{tabular}}
\caption{Inference settings. Input sizes refer to the longer side after upscaling; BRESSAY pages are about 490--540\,px wide (split medians). The native prompts are empty for LightOnOCR, ``OCR:'' for PaddleOCR-VL and ``Text Recognition:'' for GLM-OCR; the error-preserving instruction is given in Appendix~\ref{sec:prompt}.}
\label{tab:settings}
\end{table}

\section{Escriba-OCR-1B}
\label{sec:escriba}

\paragraph{Model.} We fine-tune LightOnOCR-3-1B \citep{lightonocr3blog2026}, which keeps the LightOnOCR-2 architecture \citep{lightonocr2026}: a Pixtral vision encoder (24 layers, patch size 14, $2\times2$ spatial merge) and a Qwen3 decoder (28 layers, hidden size 1024). The model transcribes a page from an image-only prompt, and we keep that interface.

\paragraph{Training data and objective.} Training examples are pages, paragraph crops and line crops of the 647 training pages, sampled with probabilities 0.4, 0.3 and 0.3. The target is the BRESSAY transcription verbatim, including errors, hyphenation and markup, followed by the end-of-turn token. The loss is token-level cross-entropy on target tokens only, summed over the micro-batches of an optimiser step and divided by their total number of target tokens, so that a short line does not weigh as much as a page; logits are computed only at target positions. Every image is resized to 1200\,px on its longer side, which yields about 1,700 visual tokens per page, and a light ``phone-photo'' augmentation (rotation $\pm1^\circ$, perspective, scaling, blur, shadow, JPEG compression) is applied.

\paragraph{Optimisation.} All parameters are trained, the vision encoder at $0.1\times$ the learning rate, with AdamW ($\beta=0.9,0.95$, no weight decay), learning rate $6\times10^{-5}$ with 3\% warm-up and cosine decay, gradient clipping at 1.0, effective batch 8 and 12,000 examples (1,500 steps). Weights are bf16 with fp32 master copies in the optimiser, since pure-bf16 updates vanish at this learning rate. Training takes 85 minutes on one A100 40\,GB (loss curve in Figure~\ref{fig:training}, appendix); the whole study, including screening runs and baselines, used about 28 GPU-hours. Table~\ref{tab:hparams} in the appendix lists all hyperparameters.

\paragraph{Implementation notes.} With transformers 5.22, the reference model class silently leaves \emph{all} decoder weights of this checkpoint randomly initialised because of a key-prefix mismatch: the base model's loss on a ground-truth page is 12.1 instead of 0.35. A save--load round trip through vLLM and an overfitting test on 16 pages expose the problem; we ran both before any experiment and recommend them when fine-tuning recent VLM checkpoints. Gradient checkpointing makes training 2.8 times slower on an NVIDIA L4, and training the vision encoder without it requires a 40\,GB GPU.

\section{Results}
\label{sec:results}

\begin{table}[t]
\centering\small
\resizebox{\textwidth}{!}{%
\begin{tabular}{lrrrrrrrrr}
\toprule
& & \multicolumn{4}{c}{Error rates (\%) $\downarrow$} & \multicolumn{3}{c}{Markup recall (\%) $\uparrow$} & \\
\cmidrule(lr){3-6}\cmidrule(lr){7-9}
System & Params & CER & CER$_{\text{case}}$ & CER$_{\text{text}}$ & WER & \mk{--t--} & \mk{--xxx--} (prec.) & \mk{\#\#t\#\#} & Lines (\%) \\
\midrule
\csname @@input\endcsname tables/main.tex
\bottomrule
\end{tabular}}
\caption{BRESSAY test set (199 pages), the fine-tuned model and the strongest open baselines (Table~\ref{tab:open} lists all 22). All CER variants are capped per page (Eq.~\ref{eq:cer}). Precision for \mk{--xxx--} is in parentheses; ``--'' means the system never produced the markup. $^\dagger$Evaluated on ground-truth line crops, so its line structure is correct by construction.}
\label{tab:main}
\end{table}

Table~\ref{tab:main} and Figure~\ref{fig:overview} summarise the test results, and Table~\ref{tab:boot} gives bootstrap intervals. Escriba-OCR-1B reaches 2.06\% CER, against 5.60\% for the best zero-shot baseline, Qwen3-VL 235B-A22B ($\Delta=-3.54$, $[-4.19,-2.94]$), and 8--10\% for the other strong baselines. The gap is similar on the text alone (1.63\% against 4.61\%), so it does not come from markup. Two further training runs with different seeds reach 2.05\% and 2.03\%; over the three seeds, CER is $2.05 \pm 0.02$, CER$_{\text{text}}$ $1.64 \pm 0.02$ and error preservation $24.9 \pm 1.3$ (mean $\pm$ SD). The released model is thus typical of the recipe, and the spread across seeds is about a hundredth of the gap to the best baseline. Markup recall varies by at most 2 points across seeds (Table~\ref{tab:seeds} in the appendix). Fine-tuning also teaches the transcription conventions that zero-shot models cannot follow: Escriba recovers 38\% of legible deletions, 77\% of illegible deletions (with 68\% precision) and 67\% of insertions, whereas no zero-shot model recovers more than 16\% of insertions (Qwen3.5 27B) or any deletions with useful precision. Escriba also reproduces the line structure of 97\% of the pages.

\begin{table}[t]
\centering\small
\begin{tabular}{lrrr}
\toprule
Escriba-OCR-1B minus \dots & $\Delta$ CER [95\% CI] & $\Delta$ CER$_{\text{text}}$ [95\% CI] & $\Delta$ Pres. [95\% CI] \\
\midrule
Qwen3-VL 235B-A22B & $-3.54$ [$-4.19$, $-2.94$] & $-2.98$ [$-3.58$, $-2.44$] & $+2.2$ [$-0.8$, $+5.1$] \\
Qwen3.5 27B & $-5.95$ [$-8.38$, $-4.05$] & $-5.03$ [$-7.14$, $-3.37$] & $+5.2$ [$+1.9$, $+8.4$] \\
LightOnOCR-3 4B & $-6.38$ [$-7.49$, $-5.43$] & $-5.51$ [$-6.61$, $-4.55$] & $+8.7$ [$+5.7$, $+11.8$] \\
Qwen2.5-VL 72B & $-6.45$ [$-7.77$, $-5.35$] & $-5.95$ [$-7.26$, $-4.90$] & $+1.4$ [$-2.6$, $+5.4$] \\
LightOnOCR-3 1B (base) & $-6.91$ [$-7.64$, $-6.20$] & $-6.11$ [$-6.82$, $-5.39$] & $+10.7$ [$+7.4$, $+13.9$] \\
HTR-Flor & $-8.27$ [$-9.22$, $-7.34$] & $-7.85$ [$-8.75$, $-6.98$] & $+4.4$ [$-1.4$, $+10.1$] \\
\bottomrule
\end{tabular}
\caption{Paired page-level bootstrap on the test set ($B=2{,}000$), in percentage points. Negative $\Delta$ CER and positive $\Delta$ Pres.\ favour Escriba. The intervals reflect variation over pages, not over training seeds; across three seeds the CER of Escriba varies by $\pm 0.02$ points.}
\label{tab:boot}
\end{table}

\begin{table}[t]
\centering\small
\begin{tabular}{lrrrrrrr}
\toprule
Model & Params & CER & WER & Pres. & Ins. rec. & Lines & Loops \\
\midrule
\csname @@input\endcsname tables/open.tex
\bottomrule
\end{tabular}
\caption{All 22 open baselines on the BRESSAY test set (\%, except Loops), sorted by CER. Pres.: error preservation; Ins. rec.: insertion recall; Loops: pages whose generation hit the token limit, which was 3,000 tokens for OpenRouter models and 2,048 for local ones. MoE models list total (active) parameters. WER is not capped, so looping models can exceed 100\%. $^\dagger$Trained on BRESSAY and evaluated on ground-truth line crops, so its line structure is correct by construction. For comparison, Escriba-OCR-1B: 2.06 / 5.9 / 23.6 / 67 / 97 / 0.}
\label{tab:open}
\end{table}

Table~\ref{tab:open} lists all 22 open baselines. Among the general-purpose VLMs, the Qwen family leads at every size, while Gemma 3 and Llama 4 Scout frequently stop transcribing and start paraphrasing, or loop until the token limit; Gemma 3 4B loops on 125 of the 199 pages. Only Qwen3-VL 235B (79\%), Gemma 4 31B (70\%) and Qwen3.5 27B (53\%) get the line structure right on more than half of the pages. The OCR-specialised LightOnOCR-3 1B and 4B are as accurate as general VLMs 7 to 70 times larger, which motivated our choice of base model.

\begin{table}[t]
\centering\small
\begin{tabular}{llrr}
\toprule
System & Source & CER (\%) & WER (\%) \\
\midrule
LITIS & ICDAR 2024 competition & 8.80 & 17.66 \\
Demokritos & ICDAR 2024 competition & 6.62 & 19.49 \\
PERO (competition winner) & ICDAR 2024 competition & 3.77 & 10.08 \\
DAN & \citet{metadan2025} & 2.43 & 7.85 \\
Meta-DAN & \citet{metadan2025} & 2.38 & 8.46 \\
Faster DAN & \citet{metadan2025} & 2.34 & 8.03 \\
MT-DAN & \citet{metadan2025} & 2.33 & 8.28 \\
\midrule
Escriba-OCR-1B & this work & \textbf{2.14} & \textbf{6.24} \\
\bottomrule
\end{tabular}
\caption{The best published page-level results on the BRESSAY test set (199 pages), case-sensitive with edit markup and without a per-page cap. Competition results from \citet{neto2024competition}, with team names and the LITIS and Demokritos results from \citet{vossos2024thesis}; DAN-family results are the main results of \citet{metadan2025} with the FCN-M encoder (an ablation setting of MT-DAN reaches 2.24\% CER). Escriba's CER ranges from 1.99\% to 2.14\% across the readings of the competition's scoring that we computed (Section~\ref{sec:results}); the table gives the highest.}
\label{tab:published}
\end{table}

\paragraph{Comparison with published BRESSAY results.}
Published results on BRESSAY use the metric of the ICDAR 2024 competition \citep{neto2024competition}: CER and WER are case-sensitive, the edit markup is part of the text, and pages are not capped. The competition's scoring script is not public, so we score Escriba under several readings of it: line breaks kept or replaced by spaces, errors micro-averaged or averaged over pages, and markup kept verbatim or, with line breaks kept, mapped to single symbols as in the BRESSAY authors' toolkit. Escriba reaches 2.14\% CER and 6.24\% WER with line breaks kept and micro-averaging, and 1.99--2.14\% CER across these variants; over the three seeds, CER is $2.12 \pm 0.02$ and WER $6.20 \pm 0.04$. This is below all published page-level results we know of (Table~\ref{tab:published}), including the DAN family, whose main results reach 2.33\% and whose best ablation setting reaches 2.24\%, and its WER is about 1.6 points lower than the best published one. The comparison is not fully controlled: the other systems were trained by their authors under their own settings, and the scoring details may differ slightly. At line level, where systems receive ground-truth line crops, the competition winner reached 2.88\% CER.

\paragraph{Efficiency.} Escriba transcribes a page in 2.3\,s on an A100 when pages are processed one at a time and at 1.6 pages per second in a vLLM batch, or about \$0.35 per 1,000 pages at an on-demand price of \$1.99 per GPU-hour. The strongest zero-shot baselines served through OpenRouter took 14--16\,s per page, including network and provider queueing, and cost \$1.8--3.2 per 1,000 pages there; served locally, the largest of them would need one or several 80\,GB GPUs.

\paragraph{CPU inference.} We converted the model to GGUF for llama.cpp and quantised the language model and the vision projector separately. On 30 test pages drawn at random, the mean page CER of every variant is within 0.05 points of the unquantised model served by vLLM (2.54--2.60\% against 2.56\%; Table~\ref{tab:cpu}), well within page-to-page noise. CER on the text alone is about 0.2 points higher for every llama.cpp variant, F16 included (2.04--2.09\% against 1.85\%), so this small gap comes from the runtime rather than from quantisation. On a 4-vCPU cloud VM (AMD EPYC Turin, 16\,GB RAM), the 4-bit language model with a 4-bit vision projector, 0.66\,GB in total, transcribes a page in a median of 53\,s with under 3\,GB of RAM, provided that the server's prompt cache is disabled (\texttt{--cache-ram 0}); otherwise memory grows by about 0.3\,GB per page. Two thirds of that time (35\,s) is spent encoding the roughly 1,700 visual tokens, and the text is then generated at 45 tokens per second. Encoding time depends mainly on the projector and generation speed on the language model; on this CPU the 8-bit projector encodes the image more slowly than the 16-bit one (48 against 37\,s). A class set of 40 essays takes about 35 minutes on an ordinary computer, without a GPU and without sending the essays anywhere.

\begin{table}[t]
\centering\small
\begin{tabular}{lrrrrrr}
\toprule
Language model + projector & Size & CER (\%) & Image & Text & Time per page & Peak RAM \\
\midrule
bf16, vLLM on A100 (reference) & 2.0\,GB & 2.56 & -- & -- & 2.3\,s & -- \\
F16 + F16 & 2.0\,GB & 2.60 & 41\,s & 28\,tok/s & 69\,s & 3.6\,GB \\
Q8\_0 + Q8\_0 & 1.1\,GB & 2.60 & 49\,s & 38\,tok/s & 70\,s & 2.9\,GB \\
Q4\_K\_M + F16 & 1.2\,GB & 2.55 & 37\,s & 47\,tok/s & 55\,s & 3.0\,GB \\
Q4\_K\_M + Q8\_0 & 0.85\,GB & 2.54 & 48\,s & 45\,tok/s & 66\,s & 2.8\,GB \\
Q4\_K\_M + Q4\_K\_M & 0.66\,GB & 2.60 & 35\,s & 45\,tok/s & 53\,s & 2.7\,GB \\
\bottomrule
\end{tabular}
\caption{GGUF variants in llama.cpp (commit \texttt{b019be0}) on one 4-vCPU VM (AMD EPYC Turin, 16\,GB RAM), 30 random test pages processed one at a time with the prompt cache disabled. CER is the mean of capped page CER. Image is the median time to encode the page image (about 1,700 visual tokens), Text the median generation speed, and Time per page the median end-to-end latency; Peak RAM is the server's peak resident memory over the 30 pages.}
\label{tab:cpu}
\end{table}

\section{Analysis}
\label{sec:analysis}

\subsection{Where the remaining errors are}

Table~\ref{tab:decomp} decomposes the character errors. Accents contribute little: stripping all diacritics lowers CER by only 0.07--0.29 points for every system. Punctuation contributes more, 0.3 points for Escriba and 1.1--1.5 points for the strongest zero-shot models. The edit counts show the main difference between the fine-tuned model and the baselines: Escriba's errors are mostly letter substitutions, with insertions and deletions of similar size, while the zero-shot models make as many deletions as substitutions, because they drop words and whole lines. Escriba makes 2.2 times fewer substitutions and 5 times fewer deletions than Qwen3-VL 235B.

\begin{table}[t]
\centering\small
\begin{tabular}{lrrrrrrr}
\toprule
& \multicolumn{4}{c}{CER (\%)} & \multicolumn{3}{c}{Character edits} \\
\cmidrule(lr){2-5}\cmidrule(lr){6-8}
System & CER & with case & w/o accents & w/o punct. & Subst. & Ins. & Del. \\
\midrule
Escriba-OCR-1B & 2.06 & 2.13 & 1.99 & 1.77 & 5,877 & 2,489 & 2,593 \\
Qwen3-VL 235B-A22B & 5.60 & 6.44 & 5.45 & 4.49 & 12,999 & 3,690 & 13,076 \\
Qwen3.5 27B & 8.01 & 8.40 & 7.90 & 6.61 & 15,636 & 6,276 & 13,464 \\
LightOnOCR-3 4B & 8.44 & 8.52 & 8.30 & 6.99 & 17,279 & 10,598 & 18,625 \\
Qwen2.5-VL 72B & 8.51 & 9.12 & 8.22 & 7.17 & 20,278 & 4,187 & 17,796 \\
LightOnOCR-3 1B (base) & 8.97 & 9.83 & 8.76 & 7.56 & 23,308 & 5,422 & 18,928 \\
HTR-Flor$^\dagger$ & 10.33 & 10.43 & 10.15 & 9.02 & 28,434 & 3,705 & 22,724 \\
\bottomrule
\end{tabular}
\caption{Decomposition of test errors (all CER values capped per page). ``w/o accents'' and ``w/o punct.'' recompute CER after stripping diacritics or punctuation from both sides. Edit counts are summed over the 531k reference characters (case-insensitive). $^\dagger$Ground-truth line crops.}
\label{tab:decomp}
\end{table}

\subsection{Per-page behaviour}

\begin{figure}[t]
\centering
\includegraphics[width=\textwidth]{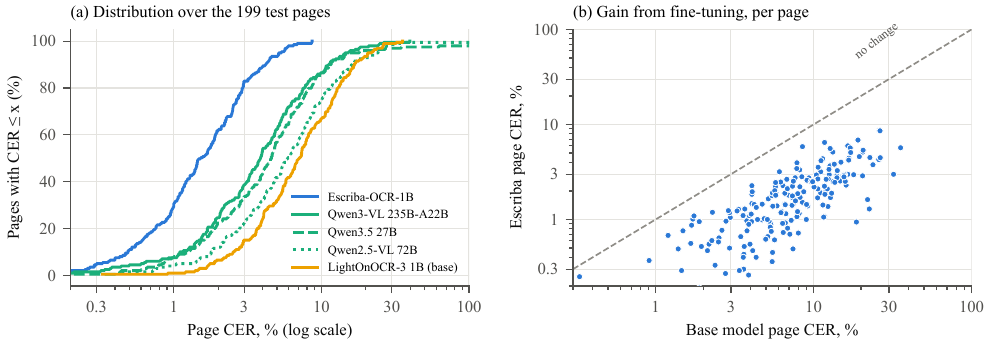}
\caption{(a) Cumulative distribution of page CER on the test set. (b) Page CER of Escriba against that of its base model. Every point lies below the diagonal: fine-tuning improves every page.}
\label{fig:perpage}
\end{figure}

Figure~\ref{fig:perpage}a shows the distribution of page CER. Escriba transcribes 30\% of the pages with less than 1\% CER and 83\% with less than 3\%, with a median of 1.52\% and a worst page of 8.7\%. The best zero-shot model has a median of 3.87\% and a long tail up to 39.8\%, and the 27B and 72B models fail completely on some pages. Compared page by page with Qwen3-VL 235B, Escriba is better on 170 pages, worse on 5 and within 0.5 points on 24. Fine-tuning reduces CER on every one of the 199 pages relative to the base model (Figure~\ref{fig:perpage}b), and the gain is largest on the pages the base model found hardest.

\subsection{What models do with misspellings}

\begin{figure}[t]
\centering
\includegraphics[width=\textwidth]{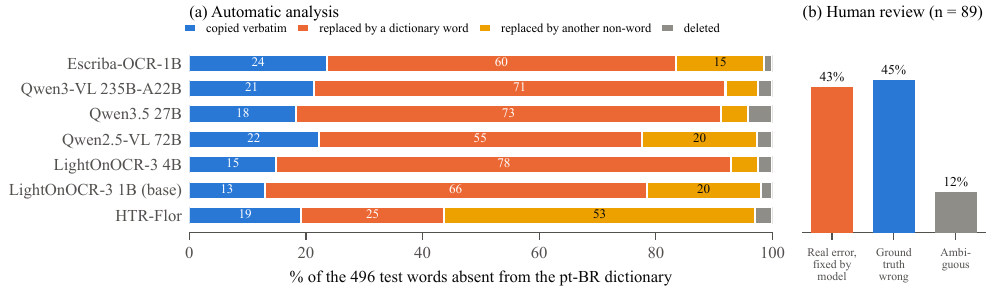}
\caption{(a) Output for the 496 test words that are absent from the pt-BR dictionary. (b) A teacher's verdict on 89 judgeable cases where Escriba replaced such a word with a dictionary word.}
\label{fig:errors}
\end{figure}

Figure~\ref{fig:errors}a shows what each system outputs for the 496 likely misspellings. All 21 zero-shot VLMs replace 39--78\% of them with a dictionary word. The strongest readers tend to sit at the top of that range (Qwen3-VL 235B normalises 71\%, Qwen3.5 27B 73\% and LightOnOCR-3 4B 78\%), and the lowest rates mostly belong to weak readers, which misread rather than normalise; Qwen2.5-VL 72B, fourth on CER, is an exception at 55\%. Fine-tuning raises Escriba's verbatim copies from 13\% for the base model to 24\% ($\Delta=+10.7$, $[+7.4,+13.9]$) and lowers its invented errors from 3.7\% to 1.4\%, but Escriba still replaces 60\% of the likely misspellings with dictionary words. Against the best zero-shot readers, its preservation advantage is small and partly within noise (+2.2 points against Qwen3-VL 235B, $[-0.8,+5.1]$), and several of them invent fewer errors: 0.7\% for Qwen3-VL 235B and 0.4\% for Qwen3.5 27B and LightOnOCR-3 4B. HTR-Flor behaves differently: without a strong language model it replaces only 25\% of the likely misspellings with dictionary words, but it misreads 53\% of them and turns 24\% of correctly spelled words into non-words. A strong language prior thus improves reading and erases errors, while a weak one keeps errors and invents new ones. Figure~\ref{fig:examples} shows typical cases.

\begin{figure}[t]
\centering
\begin{minipage}{\textwidth}
\small
\csname @@input\endcsname tables/examples.tex
\end{minipage}
\caption{Test lines (image upscaled $4\times$) with the reference and three transcriptions; differences from the reference are highlighted and \textcolor{red!70!black}{\textbf{\textvisiblespace}} marks deleted text. (a) Escriba reproduces the illegible deletion, the insertion and the legible deletion, while both zero-shot models drop the markup. (b) Escriba keeps the misspelling \emph{esteriotipando}, which both zero-shot models correct to \emph{estereotipando}. (c) A failure case, representative of the 60\% of likely misspellings that Escriba still normalises: all three systems correct \emph{entreternimento} to \emph{entretenimento}; our human review (§\ref{sec:review}) confirmed that the student wrote it. Qwen2.5-VL writes these pages without the original line breaks, so its closest span is shown.}
\label{fig:examples}
\end{figure}

\subsection{Human review of apparent corrections}
\label{sec:review}

A teacher, the author, reviewed a random sample of 96 test cases in which Escriba wrote a dictionary word in place of a non-dictionary reference word. For each case the review tool showed the line image, the reference, Escriba's output and, for context, a second system's output, and the reviewer decided whether the student had really written the misspelling. The review was not blind to which output came from Escriba. Seven cases could not be judged, because the tool showed only the first half of a word hyphenated across lines. Seven others had been answered as ``other'' with a free-text note and were assigned to a category from that note.

Of the 89 judgeable cases (Figure~\ref{fig:errors}b), 38 (43\%, Wilson 95\% CI 33--53\%) were genuine misspellings normalised by the model. Another 40 (45\%, 35--55\%) were errors in the BRESSAY transcription: the student had written the correct form, and in 38 of these the model's output matched it, while in the other two the model was wrong as well (e.g.\ \emph{catástrofe} read as \emph{amnistia}). The remaining 11 (12\%, 7--21\%) were ambiguous.

Dictionary-based preservation therefore likely understates every system, although we reviewed only Escriba's outputs. For Escriba, $|S|=496$ contains about $297 \times 40/89 \approx 134$ ground-truth errors among the words it ``corrected''. Removing them from $S$ raises its preservation from 117/496 (24\%) to $117/362 \approx 32\%$ (30--35\% over the confidence interval of the 45\% share), or to 36\% if the ambiguous cases are removed as well. We release the 40 confirmed corrections as an erratum to BRESSAY.

\subsection{Remedies for normalisation that did not work}

In the spirit of contrastive decoding \citep{li2023contrastive,leng2024vcd}, we penalised tokens that the decoder favours without the image, scoring each token with
\begin{equation}
s(t) = \log p(t \mid I, y_{<k}) - \alpha \log p(t \mid y_{<k}) \quad \text{subject to } p(t\mid I,y_{<k}) \geq 0.1 \max_{t'} p(t'\mid I,y_{<k}).
\end{equation}
With $\alpha=0.5$, preservation on 50 validation pages moved from 23.1\% to 22.3\%, invented errors rose from 2.2\% to 2.8\% and decoding was twice as slow, in line with findings that decoding-time interventions do not reliably restore visual grounding \citep{karamolegkou2026readingorguessing}. Sampling training crops that contain a likely misspelling four times as often changed preservation by $+2$ points, within noise, and 1540\,px input by $+1.3$ points. Together with the human review, these negative results suggest that a sizeable part of the residual normalisation is visual rather than a pure language-model prior: at about 19\,px per line, accents, doubled letters and letter pairs such as \emph{n}/\emph{u} are often genuinely ambiguous.

\subsection{Ablations, data scale and robustness}

Table~\ref{tab:ablation} summarises the screening study on the validation split. Training the vision encoder is the largest single gain: $-1.9$ and $-2.0$ points at matched learning rates of $3\times10^{-5}$ and $6\times10^{-5}$, and still $-1.6$ points ($[-1.9,-1.3]$) against the best decoder-only run, which used the higher rate. This is consistent with the domain shift from printed documents to small handwriting. Raising the learning rate from $1\times10^{-5}$ to $3\times10^{-5}$ helped the decoder-only runs; further increases changed CER by less than the noise. Line crops, photo augmentation on clean scans, 1540\,px input and misspelling over-sampling all changed CER by less than the seed-to-seed variation, and repetition penalties of 1.0 and 1.05 gave identical results.

\begin{table}[t]
\centering\small
\begin{tabular}{lcccc}
\toprule
Configuration (validation, 154 pages) & Examples & CER & Pres. & Ins. rec. \\
\midrule
Base model, zero-shot & -- & 10.01 & 14.1 & 0 \\
Decoder only, lr $1\times10^{-5}$ & 2.4k & 6.12 & 19.6 & 49 \\
Decoder only, lr $3\times10^{-5}$ (seed 0 / seed 1) & 2.4k & 4.70 / 5.33 & 21.5 / 20.7 & 57 / 53 \\
Decoder only, lr $6\times10^{-5}$ & 2.4k & 4.44 & 21.7 & 57 \\
Full model (vision at $0.1\times$ lr), lr $3\times10^{-5}$ & 2.4k & 2.81 & 25.7 & 64 \\
\quad pages + paragraphs only (no line crops) & 2.4k & 3.39 & 25.4 & 63 \\
\quad + photo augmentation & 2.4k & 2.73 & 25.7 & 63 \\
Full model, lr $6\times10^{-5}$ & 2.4k & 2.46 & 26.7 & 68 \\
\quad 1540\,px input instead of 1200\,px & 2.4k & 2.23 & 28.0 & 71 \\
\quad no augmentation & 8k & 2.25 & 26.7 & 64 \\
\quad misspelling over-sampling ($4\times$), augmentation & 8k & 2.15 & 26.2 & 70 \\
\quad augmentation, checkpoint at 8k of the 12k run & 8k & 2.14 & 24.3 & 68 \\
\quad \textbf{augmentation (final model)} & \textbf{12k} & \textbf{2.03} & 25.9 & 69 \\
\bottomrule
\end{tabular}
\caption{Screening on the validation split (CER capped per page, error preservation and insertion recall, \%). Between decoder-only runs, seed-to-seed variation is about 0.6 CER points and evaluation variation about 0.5, so differences below about 1 point are inconclusive; full-model runs vary much less (test SD 0.02 over three seeds).}
\label{tab:ablation}
\end{table}

\begin{figure}[t]
\centering
\includegraphics[width=\textwidth]{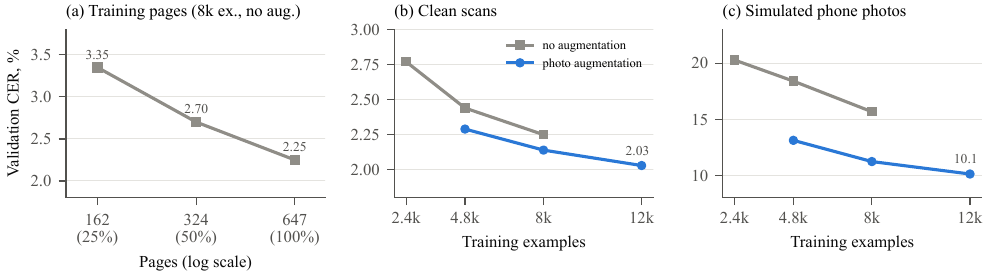}
\caption{Validation CER (capped) as a function of (a) the number of training pages, with 8k examples and no augmentation, and (b, c) training examples with and without photo augmentation, on clean scans and on simulated phone photos.}
\label{fig:scaling}
\end{figure}

\paragraph{Data scale.} With a fixed budget of 8,000 examples and no augmentation, training on 25\%, 50\% and 100\% of the training pages gives 3.35\%, 2.70\% and 2.25\% validation CER (Figure~\ref{fig:scaling}a), $-0.65$ and then $-0.45$ points per doubling: diminishing returns, but not yet flat. A few hundred additional transcribed pages would likely still pay off.

\paragraph{Training length and robustness.} Validation CER keeps falling up to 12k training examples (Figure~\ref{fig:scaling}b), although the last step, from 2.14\% at 8k to 2.03\% at 12k examples, is small relative to the variation we can resolve. On synthetically degraded validation pages, photo augmentation lowers CER at a matched 8k examples from 15.7\% to 11.2\%, and to 10.1\% after 12k examples (Figure~\ref{fig:scaling}c), against 30.9\% for the base model. The degraded pages use the same family of transformations as the augmentation, so this measures robustness to the simulation, not to real photos.

\begin{table}[t]
\centering\small
\begin{tabular}{lcccccc}
\toprule
& \multicolumn{4}{c}{BRESSAY} & \multicolumn{2}{c}{BFL (40 letters)} \\
\cmidrule(lr){2-5}\cmidrule(lr){6-7}
$\alpha$ & Val.\ CER & Test CER & Del.\ rec. & Lines & CER & Unseen words \\
\midrule
0 (Escriba) & 2.05 & 2.09 & 77.6 & 97.5 & 5.12 & 39.6 \\
0.1 & 1.98 & 1.99 & 70.3 & 97.5 & 4.70 & 36.5 \\
\textbf{0.2 (selected)} & 1.99 & 2.00 & 61.4 & 93.5 & 4.46 & 34.6 \\
0.3 & 2.16 & 2.11 & 50.6 & 93.0 & 4.34 & 33.2 \\
0.4 & 2.39 & 2.42 & 37.8 & 85.9 & 4.20 & 31.4 \\
0.5 & 2.97 & 3.14 & 16.2 & 76.4 & 4.20 & 31.5 \\
0.7 & 4.48 & 4.40 & 0.4 & 73.9 & 5.22 & 30.7 \\
1 (base model) & 10.04 & 9.41 & 0 & 14.6 & 34.32 & 31.4 \\
\bottomrule
\end{tabular}
\caption{Interpolating Escriba's weights with its base model's (WiSE-FT, $\alpha$ is the base model's share). CER is capped per page (\%); \emph{Del.\ rec.} is the recall of illegible deletions on the test set and \emph{Lines} the share of test pages with the right number of lines; \emph{Unseen words} is the share of BFL word tokens absent from the BRESSAY training transcriptions that are misread. All models use Escriba's interface (image only, 1200\,px), so the base-model row differs from the 1600\,px results in the text, and BFL CER includes the comments that the models append (on 1 letter at $\alpha = 0.7$ and 36 at $\alpha = 1$). $\alpha$ was selected on the validation split only.}
\label{tab:wise}
\end{table}

\subsection{Out-of-domain test and weight interpolation}
\label{sec:ood}

\paragraph{Test set.} To test Escriba outside BRESSAY, we use the Brazilian Forensic Letter database (BFL) \citep{freitas2008bfl}, in which 315 writers each copied the same letter, about 130 words long, three times. The letters were scanned at 300\,dpi in grey levels, and the text differs from an essay in genre and vocabulary: it is a job application written in the first person, with an address block, proper nouns and abbreviations such as \emph{V.\,Sas.} and \emph{S.\,A.}, and many writers use block capitals. We drew one letter from each of 40 writers at random. The author transcribed them in two passes. In the first, he started from the printed text of the letter and corrected it wherever the writer deviated from it, before any model was run; starting from the printed text anchors the reference to it, so deviations he overlooked remain as model errors. In the second, he re-examined 191 points: the 183 at which Escriba and its base model agreed against the reference, and the 8 spans that Escriba marked as deleted or illegible; the review tool showed the competing readings without saying which model produced them. The second pass corrects the reference only where a model drew attention to it, which favours the models over an independent transcription, and its markup points favour Escriba; we therefore report both passes. Markup recall cannot be measured on this set, because every marked span in the reference was first proposed by Escriba. Both models receive the image alone; Escriba reads it at 1200\,px and the base model at 1600\,px, as on BRESSAY. We undo line-end hyphenation in the outputs, since the references have no line breaks, and treat \emph{S.\,A} and \emph{S.A} as the same, since the gap cannot be judged in handwriting; CER is case-insensitive and capped per page, as on BRESSAY.

\paragraph{Results.} Escriba reaches 5.1\% CER against 27.0\% for the base model (6.0\% and 27.7\% against the first-pass references). Almost all of the base model's error, however, comes from a comment in English that it appends to 31 of the 40 letters (``Note: The handwritten text has been\ldots''), which it does on only 1 of the 199 BRESSAY test pages. With the comments removed, a correction we decided on after seeing the outputs and which favours the base model, the base model reaches 4.9\%, and the difference from Escriba, $+0.2$ points ($[-0.7,+0.9]$), is within noise: Escriba is better on 17 letters and worse on 21. Against the first-pass references the difference is $+0.35$ ($[-0.5,+1.1]$). Escriba appends nothing. Of the eight spans it marks, the review confirmed six deletions on four letters, among them a struck-out \emph{portanto} reproduced verbatim and one deletion whose struck word Escriba misread; the other two are spurious illegible marks (\mk{@@???@@}).

Escriba's errors are not random. It writes \emph{sobre} for \emph{Soube} on 34 of the 40 letters and never \emph{soube}, \emph{horário} for the proper noun \emph{Horácio} on 26 and \emph{colocou-me} for \emph{coloco-me} on 9, and on one clearly written letter in block capitals it turns \emph{cordiais saudações} into \emph{comidas saudáveis}. Of the 124 word tokens of the letter (numbers excluded), 36 never occur in the BRESSAY training transcriptions: proper nouns, the postal code and abbreviations, but also first-person verb forms such as \emph{sou}, \emph{possuo} and \emph{venho}, which argumentative essays avoid. On the words that do occur in BRESSAY, Escriba misreads 6.6\% of the word tokens against 8.9\% for the base model; on the 1,481 tokens of words that never occur in it, Escriba misreads 39.6\% against 29.8\%. Fine-tuning on a single genre improves reading of its vocabulary and narrows the language prior elsewhere, the same mechanism that makes it normalise misspellings to dictionary words (§\ref{sec:review}).

\paragraph{Weight interpolation.} WiSE-FT \citep{wortsman2022wiseft} restores robustness that fine-tuning removes by interpolating the weights of the fine-tuned and the original model, $\theta(\alpha) = (1-\alpha)\,\theta_{\text{Escriba}} + \alpha\,\theta_{\text{base}}$; note that our $\alpha$ is the base model's share, the reverse of the convention in \citet{wortsman2022wiseft}. We evaluated $\alpha \in \{0, 0.1, 0.2, 0.3, 0.4, 0.5, 0.7, 1\}$ with Escriba's interface (image only, 1200\,px) on the BRESSAY validation and test splits and on BFL, and fixed the selection rule before seeing any result: the largest $\alpha$ whose validation CER is within 0.1 points of $\alpha = 0$. The rule selects $\alpha = 0.2$ (Table~\ref{tab:wise}); $\alpha = 0.3$ misses the threshold by 0.005 points, and the tolerance is smaller than the variation between vLLM evaluations (§\ref{sec:metrics}), so the choice between neighbouring values is not sharp. On the BRESSAY test set this lowers CER from 2.09\% to 2.00\% ($\Delta = -0.09$, $[-0.14,-0.04]$) with no significant change in error preservation (24.6\% against 23.4\%, $\Delta=+1.2$, $[-0.5,+3.0]$) or invented errors (1.4\%); on BFL it lowers CER from 5.1\% to 4.5\% and the error on words absent from BRESSAY from 39.6\% to 34.6\%, closing more than half of the gap to the base model at the same input size (31.4\%). The cost is the markup, which only fine-tuning teaches: in this run, recall of illegible deletions falls from 77.6\% to 61.4\% and correct line structure from 97.5\% to 93.5\% of the pages, and from $\alpha = 0.5$ on the markup collapses. Because teachers need what students struck out, we keep the released model at $\alpha = 0$. A post-hoc reading of the table, not part of the selection rule, suggests $\alpha = 0.1$ for transcribing other kinds of text: it keeps the line structure and most of the markup while improving both test sets. The $\alpha = 0$ run gives 2.09\% test CER against the 2.06\% of Table~\ref{tab:main}, within the variation between vLLM evaluations.

\section{Discussion}

A 1B model fine-tuned for 85 minutes on 647 pages reduces the error of the best zero-shot baseline, a 235B mixture-of-experts, by a factor of 2.7, and it is the only system in our study that follows the transcription conventions of the corpus. We attribute most of the gain to two ingredients, of which only the second was ablated: an OCR-specialised base model, which already reads as well as general VLMs many times its size, and training the vision encoder on the target handwriting. The result matters for schools in three ways. A 1B model runs on a single modest GPU or, quantised, on an ordinary CPU, so essays need not leave the institution; teachers care about what students struck out and added, which no zero-shot model reproduces reliably; and following a corpus's conventions, such as its casing, is what fine-tuning provides and prompting does not.

Error preservation is the open problem. Every model we tested normalises, and fine-tuning reduces this only partly; in the human review, 43\% of the words Escriba changed were genuine misspellings that it normalised. On BRESSAY, however, almost half of what looks like normalisation is noise in the reference, so error-preserving benchmarks need adjudicated, double-annotated ground truth \citep{gold2023preserving}, and preservation must be read together with the invented-error rate, without which a recogniser that misreads everything, such as HTR-Flor, looks faithful.

The out-of-domain test shows the other side of the same language prior. Fine-tuned on essays alone, Escriba reads essay vocabulary better than its base model and other vocabulary worse, so that \emph{Soube} becomes \emph{sobre} and a proper noun becomes a common word. A model meant for schools will meet letters, summaries and answers to exercises, written in the first person and full of names. Training data from more genres is the direct remedy; weight interpolation with the base model is a cheap partial one, but it trades away edit markup, which only the fine-tuning data teach.

\section{Conclusion}

We introduced a benchmark for error-preserving transcription of handwritten Brazilian Portuguese essays and evaluated 22 open baselines on it. The VLMs among them normalise students' misspellings to a large extent, and none of the baselines reproduces the edit markup reliably. Escriba-OCR-1B, a 1B model fine-tuned in 85 minutes on one GPU, reduces the character error rate of the best zero-shot baseline by a factor of 2.7, improves on the best published page-level result on BRESSAY, reproduces edit markup and runs on a single GPU or, quantised to 4 bits, on an ordinary CPU. On 40 letters from a second corpus it reads as well as its base model once that model's appended comments are removed, but misreads words absent from its training data more often, a gap that weight interpolation narrows by about half. Future work includes real classroom photos, training data from more genres, adjudicated ground truth and training signals that reward copying what is drawn over what is likely.

\section*{Limitations}

The model was trained on a single dataset, and our only out-of-domain test is small: 40 letters that copy one fixed text, with few struck-out words, transcribed by a single reviewer whose second pass examined only the points where both models agreed against the reference, plus the spans Escriba marked. BRESSAY pages were collected from online platforms, cropped and at low resolution; performance on real classroom photos is untested, and our robustness numbers rely on synthetic degradations. The BRESSAY transcriptions contain errors, and our review covers 96 cases judged by a single, non-blind annotator. The error-preservation metric is dictionary-based: it misses agreement errors (\emph{as pessoa}) and counts foreign words and proper nouns as misspellings. The bootstrap intervals do not include training variance; three seeds of the final configuration suggest that it is small, but three runs are a small sample. CPU latency was measured on a cloud VM, not on school hardware. Baselines were run under different input sizes and token limits (Table~\ref{tab:settings}), and OpenRouter served several of them from multiple providers with unknown quantisation, which may lower their scores. BRESSAY has been public since 2024, so the evaluated models may have seen its test images. The general VLMs received a single error-preserving prompt; we did not test prompt variants or few-shot examples. We fine-tuned a single base model, so we cannot say how close another fine-tuned VLM would come. The comparison with published BRESSAY results rests on our reconstruction of the competition's metric, whose scoring script is not public, and on numbers reported by other authors. Finally, the study covers open-weight models only and makes no claim about proprietary systems.

\section*{Ethics Statement}
Escriba-OCR-1B transcribes and does not assess, and it should not be used for grading or other high-stakes decisions without human review. Because BRESSAY may be used for non-commercial research and teaching only, the model is released under CC BY-NC 4.0. We encourage local deployment so that students' essays are not sent to third-party services. The human review was carried out by the author, a teacher, on public dataset images.

\section*{Data and Code Availability}
\begin{sloppypar}\raggedright
The model weights are available at \url{https://huggingface.co/tardellirs/Escriba-OCR-1B} (CC BY-NC 4.0), and GGUF files for CPU inference at \url{https://huggingface.co/tardellirs/Escriba-OCR-1B-GGUF}. An online demo is available at \url{https://huggingface.co/spaces/tardellirs/Escriba-OCR-Demo}, and all resources are collected at \url{https://huggingface.co/collections/tardellirs/escriba-ocr}. Code and data are available at \url{https://github.com/tardellirs/escriba-ocr}. The repository contains:
the benchmark code (normalisation, metrics, bootstrap, and the runners for OpenRouter and vLLM);
the fine-tuning scripts, including the checkpoint loader described in §\ref{sec:escriba};
a client, a command-line tool and a local viewer that shows each page next to its transcription and a word-level diff against the reference;
the generators for every table and figure;
the outputs of all 23 systems on the test split;
the human-review verdicts and the BRESSAY erratum;
our reference transcriptions of the 40 BFL letters (both passes), the outputs of the out-of-domain and weight-interpolation runs, and their scoring code.
The code is released under Apache~2.0; model outputs and review data under CC BY-NC 4.0.
\end{sloppypar}

\section*{Acknowledgements}
We thank the BRESSAY authors for releasing the dataset, and LightOn for releasing LightOnOCR under an open license.

\bibliographystyle{plainnat}
\bibliography{refs}

\appendix
\section{Hyperparameters, seeds and providers}

\begin{table}[ht]
\centering\small
\begin{tabular}{ll}
\toprule
Base model & \texttt{lightonai/LightOnOCR-3-1B} (Pixtral encoder, Qwen3 decoder) \\
Trainable & all parameters; vision encoder at $0.1\times$ learning rate \\
Optimiser & AdamW, $\beta_1=0.9$, $\beta_2=0.95$, no weight decay, gradient clipping 1.0 \\
Learning rate & $6\times10^{-5}$, 3\% linear warm-up, cosine decay to 0 \\
Batch & micro-batch 1, 8 accumulation steps; 1,500 steps (12,000 examples) \\
Data mixture & pages / paragraphs / lines = 0.4 / 0.3 / 0.3 (training split only) \\
Input & longer side resized to 1200\,px (LANCZOS); maximum sequence 4,608 tokens \\
Augmentation & $p=0.5$: rotation $\pm1^\circ$, perspective 2\%, scale 0.95--1.05, blur $\le0.6$, shadow, JPEG 60--95 \\
Loss & token cross-entropy on target tokens, normalised per step by target-token count \\
Precision & bf16 weights, fp32 master weights and optimiser state; no gradient checkpointing \\
Hardware & 1$\times$ NVIDIA A100 40\,GB, 85 minutes; peak memory 34\,GB \\
Inference & vLLM 0.34, greedy, repetition penalty 1.05, max 2,048 new tokens, image-only prompt \\
Software & transformers 5.22.1, PyTorch 2.14, vLLM 0.34 \\
\bottomrule
\end{tabular}
\caption{Training and inference configuration of Escriba-OCR-1B.}
\label{tab:hparams}
\end{table}

\begin{figure}[ht]
\centering
\includegraphics[width=0.48\textwidth]{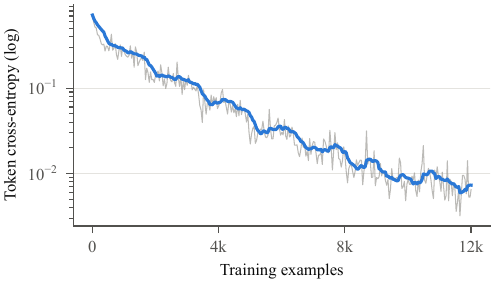}
\caption{Training loss of Escriba-OCR-1B (raw and 10-step moving average). The final loss of about 0.005 shows that the model nearly memorises the 647 training pages; validation CER nevertheless keeps improving (Figure~\ref{fig:scaling}).}
\label{fig:training}
\end{figure}

\begin{table}[ht]
\centering\small
\begin{tabular}{lrrrr}
\toprule
Metric (test, \%) & Seed 0 (released) & Seed 1 & Seed 2 & Mean $\pm$ SD \\
\midrule
CER & 2.06 & 2.05 & 2.03 & $2.05 \pm 0.02$ \\
CER$_{\text{case}}$ & 2.13 & 2.11 & 2.10 & $2.11 \pm 0.02$ \\
CER$_{\text{text}}$ & 1.63 & 1.67 & 1.63 & $1.64 \pm 0.02$ \\
WER & 5.88 & 5.83 & 5.83 & $5.85 \pm 0.03$ \\
Error preservation & 23.6 & 26.2 & 25.0 & $24.9 \pm 1.3$ \\
Invented errors & 1.40 & 1.37 & 1.33 & $1.37 \pm 0.04$ \\
\mk{--t--} recall & 37.8 & 38.5 & 38.5 & $38.3 \pm 0.4$ \\
\mk{--xxx--} recall & 77.2 & 79.2 & 79.2 & $78.5 \pm 1.1$ \\
\mk{\#\#t\#\#} recall & 67.3 & 66.8 & 67.0 & $67.0 \pm 0.3$ \\
Lines & 97.5 & 97.0 & 95.0 & $96.5 \pm 1.3$ \\
\bottomrule
\end{tabular}
\caption{Three training seeds of the final configuration on the BRESSAY test set. Only the random seed changes between runs; it controls the sampling of pages, paragraphs and lines, their order and the augmentation. The released model is seed 0.}
\label{tab:seeds}
\end{table}

\begin{table}[ht]
\centering\small
\resizebox{\textwidth}{!}{%
\begin{tabular}{ll}
\toprule
Model & Providers (number of the 199 test pages) \\
\midrule
Qwen3-VL 235B-A22B & Parasail 121, Venice 78 \\
Qwen3.5 27B & Chutes 91, DeepInfra 57, AtlasCloud 26, Alibaba 25 \\
Qwen3.5 9B & PrimeIntellect 87, Parasail 70, Together 39, DeepInfra 3 \\
Gemma 4 31B & NextBit 80, KT 75, CoreWeave 39, DeepInfra 4, ModelRun 1 \\
Gemma 3 27B & Nebius 81, Parasail 53, DeepInfra 46, NextBit 19 \\
Gemma 3 12B & DeepInfra 166, NextBit 33 \\
Gemma 3 4B & Parasail 166, DeepInfra 33 \\
Qwen2.5-VL 72B, Qwen3-VL 8B, Llama 4 Maverick & Parasail 199 \\
Qwen3-VL 30B-A3B, Mistral Small 3.2 & DeepInfra 198, other 1 \\
Llama 4 Scout & DeepInfra 199 \\
\bottomrule
\end{tabular}}
\caption{OpenRouter providers that served each model during the test run.}
\label{tab:providers}
\end{table}

\section{Error-preserving prompt}
\label{sec:prompt}
The 16 general VLMs received the page image with the following instruction (verbatim, in Portuguese). It asks for a diplomatic transcription that keeps all student errors, one output line per handwritten line, end-of-line hyphens and the BRESSAY markup.

\noindent\begin{minipage}{\linewidth}
\begin{quote}\small\raggedright
Você é um transcritor paleográfico. Transcreva a redação manuscrita da imagem de forma DIPLOMÁTICA.\\[4pt]
Regras obrigatórias:\\
1. Copie exatamente o que o aluno escreveu. NÃO corrija ortografia, acentuação, concordância, pontuação, maiúsculas/minúsculas nem palavras repetidas. Erros do aluno devem aparecer na transcrição.\\
2. Cada linha do papel vira uma linha na saída, na mesma ordem. Mantenha hífens de fim de linha (ex.: ``con-'' numa linha e ``ceito'' na seguinte). Se o aluno quebrou a palavra sem hífen, não acrescente hífen.\\
3. Separe parágrafos com uma linha em branco.\\
4. Use estas marcações:\\
\hspace*{1.5em}\texttt{-{}-texto-{}-}\quad texto riscado mas legível\\
\hspace*{1.5em}\texttt{-{}-xxx-{}-}\quad texto riscado ilegível\\
\hspace*{1.5em}\texttt{\#\#texto\#\#}\quad texto inserido acima da linha (entrelinha superior)\\
\hspace*{1.5em}\texttt{\$\$texto\$\$}\quad texto inserido abaixo da linha\\
\hspace*{1.5em}\texttt{@@???@@}\quad trecho ilegível\\
\hspace*{1.5em}Combinações são permitidas, ex.: \texttt{\#\#-{}-texto-{}-\#\#}.\\
5. Transcreva apenas o que está visível na imagem; não complete frases cortadas.\\
6. Responda somente com a transcrição: sem título, sem comentários, sem markdown, sem cercas de código.
\end{quote}
\end{minipage}

\end{document}